\pdfoutput=1
\documentclass[11pt]{article}

\usepackage{ACL2023}

\usepackage{times}
\usepackage{latexsym}
\usepackage[most]{tcolorbox}
\usepackage{amsmath}
\usepackage{pifont}
\usepackage{multirow}
\usepackage{soul}
\usepackage{booktabs}
\usepackage{amssymb}
\usepackage{adjustbox}
\usepackage{subcaption}
\usepackage{xcolor}
\usepackage{xspace}
\usepackage{algorithm}
\usepackage{algpseudocode}
\newcommand{\stitle}[1]{\vspace{1ex} \noindent{\bf #1.}}

\newcommand{\gptthree}{\textsc{GPT-3}\xspace}
\newcommand{\gptthreepointfive}{\textsc{GPT-3.5}\xspace}
\newcommand{\gptfour}{\textsc{GPT-4}\xspace}

\newcommand{\gptfouromini}{\textsc{GPT-4o-mini}\xspace}

\newcommand{\gemma}{\textsc{Gemmma}\xspace}
\newcommand{\gemmatwo}{\textsc{Gemmma2}\xspace}
\newcommand{\llamathreepointone}{\textsc{Llama3.1}\xspace}
\newcommand{\qwentwopointfive}{\textsc{Qwen2.5}\xspace}
\newcommand{\llamatwosevenb}{\textsc{Llama2-7B}\xspace}
\newcommand{\gemmatwonineb}{\textsc{Gemma2-9B}\xspace}
\newcommand{\qwenseven}{\textsc{Qwen2.5-7B}\xspace}
\newcommand{\llamaseventy}{\textsc{Llama3.1-70B}\xspace}
\newcommand{\gemmatwotwosevenb}{\textsc{Gemma2-27B}\xspace}

\newcommand{\gpt}{\textsc{GPT}\xspace}

\usepackage[T1]{fontenc}
\usepackage[utf8]{inputenc}

\usepackage{microtype}

\usepackage{inconsolata}

\tcbset{
    userstyle/.style={
        enhanced,
        colback=white,
        colframe=black,
        colbacktitle=gray!20,
        coltitle=black,
        rounded corners,
        sharp corners=north,
        boxrule=0.5pt,
        drop shadow=black!50!white,
        attach boxed title to top left={
            xshift=-2mm,
            yshift=-2mm
        },
        boxed title style={
            rounded corners,
            size=small,
            colback=gray!20
        }
    },
    replystyleg/.style={
        enhanced,
        colback=green!15,
        colframe=black,
        colbacktitle=green!30,
        coltitle=black,
        boxrule=0.5pt,
        drop shadow=black!50!white,
        rounded corners,
        sharp corners=north,
        attach boxed title to top right={
            xshift=-2mm,
            yshift=-2mm
        },
        boxed title style={
            rounded corners,
            size=small,
            colback=green!40
        }
    },
    replystyler/.style={
        enhanced,
        colback=red!15,
        colframe=black,
        colbacktitle=red!40,
        coltitle=black,
        boxrule=0.5pt,
        drop shadow=black!50!white,
        rounded corners,
        sharp corners=north,
        attach boxed title to top right={
            xshift=-2mm,
            yshift=-2mm
        },
        boxed title style={
            rounded corners,
            size=small,
            colback=red!40
        }
    }
}

\newtcolorbox{userquery}[1][]{
    userstyle,
    title=Prompt,
    #1
}

\definecolor{darkred}{RGB}{200,0,0}
\definecolor{lightgreen}{RGB}{160,230,160}
\definecolor{lightred}{RGB}{252,231,234}
\definecolor{lightyellow}{RGB}{250,253,191}
\definecolor{lightblue}{RGB}{230,240,254}
\definecolor{lightorange}{RGB}{255,223,191}
\definecolor{white}{RGB}{255,255,255}

\newcommand\hlc[2]{\sethlcolor{#1} \hl{#2}}

\newcommand{\greentext}[1]{{\hlc{lightgreen}{#1}}}

\title{DRS: Deep Question Reformulation With Structured Output}

\author{
Zhecheng Li$^\dagger$ \ \ \ \ Yiwei Wang$^{\ddagger \mathparagraph}$ \ \ \ \ Bryan Hooi$^\|$ \ \ \ \ Yujun Cai$^\mathsection$ \ \ \ \ \\ 
\ \ \ \ \textbf{Nanyun Peng}$^\ddagger$ \ \ \ \ \textbf{Kai-Wei Chang}$^\ddagger$ \\
$^\dagger$ University of California, San Diego \quad $^\ddagger$ University of California, Los Angelas \\
$^\mathsection$ The University of Queensland \quad $^\|$ National University of Singapore \\
$^\mathparagraph$ University of California, Merced \\
\texttt{zhl186@ucsd.edu}
\\
\href{https://github.com/Lizhecheng02/DRS}{\textcolor{magenta}{\texttt{https://github.com/Lizhecheng02/DRS}}}
}

\begin{document}
\maketitle
\begin{abstract}
Question answering represents a core capability of large language models (LLMs). However, when individuals encounter unfamiliar knowledge in texts, they often formulate questions that the text itself cannot answer due to insufficient understanding of the underlying information. Recent studies reveal that while LLMs can detect unanswerable questions, they struggle to assist users in reformulating these questions. Even advanced models like \gptthreepointfive demonstrate limited effectiveness in this regard. To address this limitation, we propose DRS: \textbf{D}eep Question \textbf{R}eformulation with \textbf{S}tructured Output, a novel zero-shot method aimed at enhancing LLMs’ ability to assist users in reformulating questions to extract relevant information from new documents. DRS combines the strengths of LLMs with a DFS-based algorithm to iteratively explore potential entity combinations and constrain outputs using predefined entities. This structured approach significantly enhances the reformulation capabilities of LLMs. Comprehensive experimental evaluations demonstrate that DRS improves the reformulation accuracy of \gptthreepointfive from $23.03\%$ to $70.42\%$, while also enhancing the performance of open-source models, such as \gemmatwonineb, from $26.35\%$ to $56.75\%$. 
% These results highlight the effectiveness and versatility of our method in advancing the utility of LLMs for question reformulation tasks.
\end{abstract}

\section{Introduction}
Question answering has emerged as a fundamental capability of large language models (LLMs), with recent advances from \gptthree and InstructGPT to \gptfour~\citep{gpt3, instructgpt, gpt4} demonstrating remarkable improvements on various benchmarks~\citep{qa1, qasurvey, transformerqa}. However, when humans encounter unfamiliar knowledge domains, they frequently pose questions that cannot be directly answered from the available text. Recent studies indicate that over 30\% of questions in real-world scenarios fall into this category, significantly impacting information access and learning efficiency~\citep{gaoerrorrate, yuerrorrate}. More importantly, industrial experiments have shown that effectively reformulating such questions can dramatically improve user experience and task completion rates in virtual assistant systems, with potential impact on millions of users~\citep{importantevidence}.

\begin{figure}[!tb]
    \centering
    \includegraphics[width=1.00\linewidth]{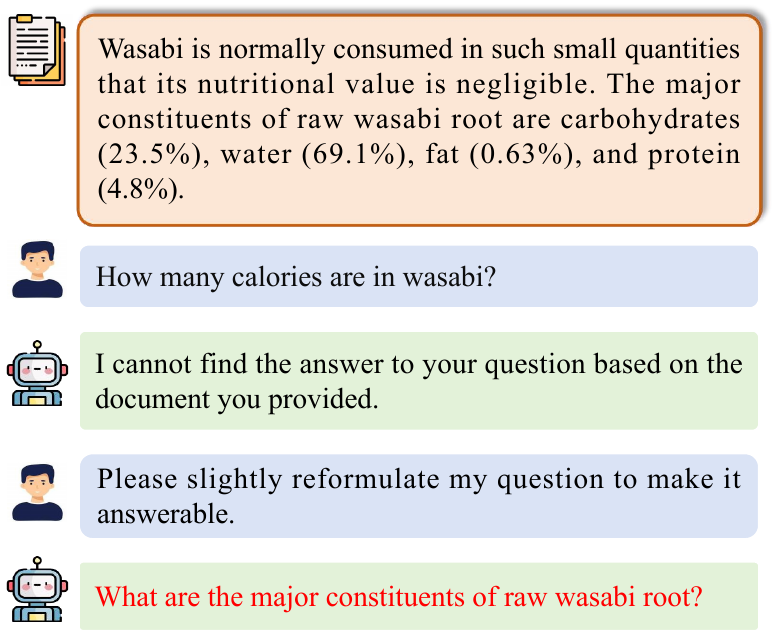}
    \caption{An example of question reformulation using large language models.
    \label{fig:example}}
    \vspace{-2mm}
\end{figure}

A successful question reformulation must satisfy two key criteria: (1) the reformulated question should be answerable based on the given text, and (2) it should preserve the core entities and intent of the original question, ensuring users obtain the information they actually seek. Consider the following example shown in Figure \ref{fig:example}: when presented with a text stating `Wasabi is normally consumed in such small quantities that its nutritional value is negligible. The major constituents of raw wasabi root are carbohydrates ($23.5\%$), water ($69.1\%$), fat ($0.63\%$), and protein ($4.8\%$)', users might ask `How many calories are in wasabi?'. While this question cannot be directly answered, a well-reformulated version would be `What are the major constituents of raw wasabi root?', which maintains the user's interest in wasabi's composition while being answerable from the text.

Previous approaches to handling unanswerable questions broadly fall into three categories: (1) detection methods that focus on identifying unanswerable questions~\citep{unans1, unans3, unans4, unans5}, (2) clarification methods that seek additional information from users~\citep{unans2, kimdata}, and (3) reformulation methods that attempt to modify questions into answerable forms~\citep{couldask}. While recent advances in LLMs have improved performance in question detection and clarification, question reformulation remains challenging due to the difficulty in balancing answerability and intent preservation. Even powerful models like \gptthreepointfive achieve only $23.03\%$ accuracy in reformulation tasks, highlighting the significant room for improvement in this area.

To address these challenges, we propose DRS (\textbf{D}eep Question \textbf{R}eformulation with \textbf{S}tructured Output), a zero-shot method that enables LLMs to effectively reformulate unanswerable questions. DRS addresses the key challenges through three innovations: (1) a systematic entity-driven approach that ensures intent preservation by explicitly tracking and maintaining key entities from the original question, (2) a structured output framework that ensures answerability by constraining the generation process to incorporate specific entities and generate questions aligned with the corresponding statements derived from the source document, and (3) an efficient DFS-based search strategy, coupled with a candidate question evaluation mechanism, enhances the method's effectiveness and practicality for real-world applications. Unlike previous methods that often sacrifice one aspect for another, DRS achieves strong performance on both answerability and entity preservation simultaneously.

We conduct extensive experiments on six diverse datasets, comparing DRS with multiple baseline approaches across different types of questions and domains. The results demonstrate that DRS significantly improves reformulation accuracy across all tested LLMs.
% , with \gptthreepointfive's performance increasing from 23.03\% to 70.42\% and open-source model \gemmatwonineb improving from 26.35\% to 56.75\%. 
Additionally, we introduce a more reliable evaluation framework using \gptfouromini, replacing the previous \llamatwosevenb evaluator to ensure more accurate assessment of reformulation quality.

Our contributions in this paper are threefold: 

(i) We propose DRS, a zero-shot method that enables LLMs to effectively reformulate unanswerable questions through entity-driven search and structured outputs. 

(ii) We demonstrate through extensive experiments that DRS significantly outperforms existing methods, improving reformulation accuracy by over $100\%$ across various LLMs and datasets. 

(iii) We introduce an improved evaluation framework based on \gptfouromini, providing more reliable assessment of question reformulation quality.

\begin{figure*}[!tb]
    \centering
    \includegraphics[width=1.00\linewidth]{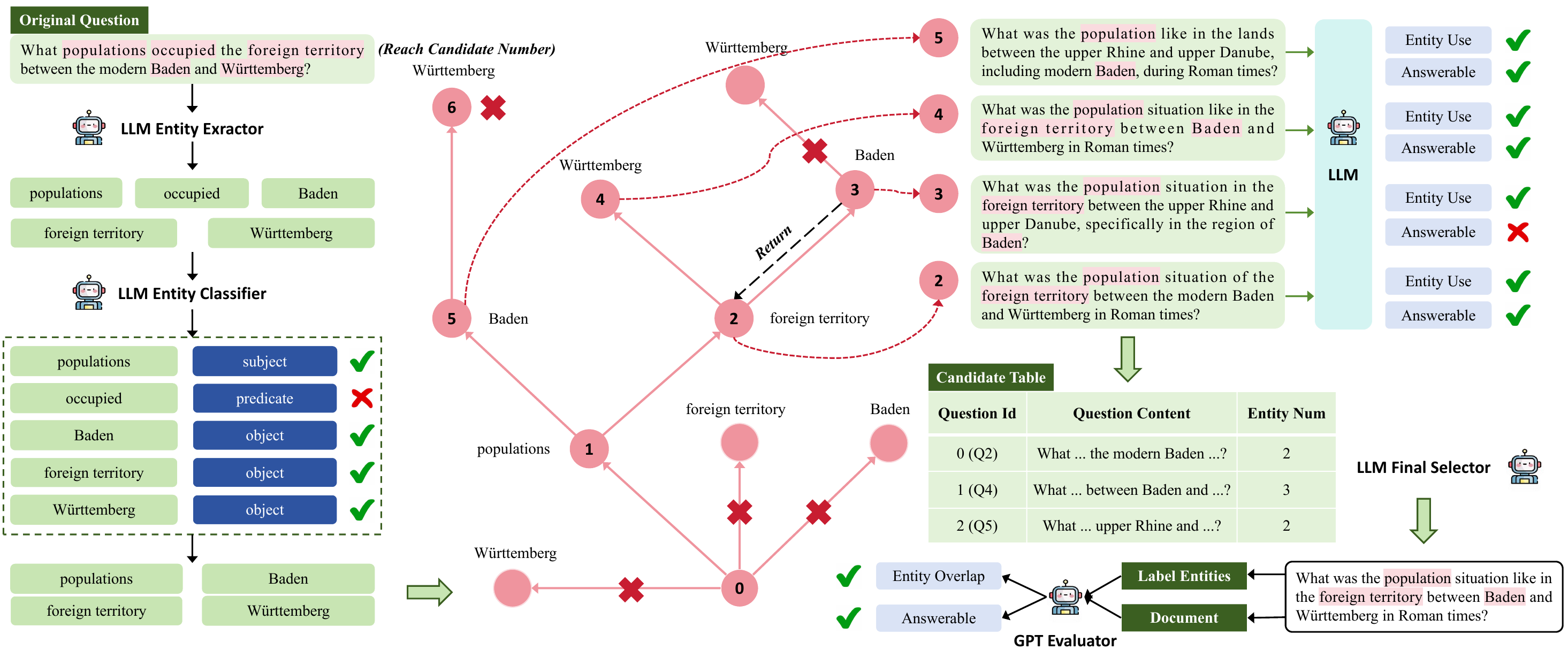}
    \caption{The complete process of our zero-shot DRS method, which mainly contains three parts - entity extraction and filtering, dfs combination search and structured question generation, and candidate question re-evaluation.
    \label{fig:algorithm}}
    \vspace{-2mm}
\end{figure*}

\section{Related Work}
Question answering has been a central focus in natural language processing (NLP)~\citep{qasurvey, bertforqa, transformerqa}, with the development of datasets like CBT and SearchQA~\citep{cbt, searchqa} designed to assess a model's ability to answer questions. However, these datasets do not address the challenge of unanswerable questions.

With the rise of \gptthree~\citep{gpt3} and even more powerful large language models (LLMs), the focus has shifted to handling unanswerable questions, which arise due to ambiguity or gaps in knowledge~\citep{unans2, gaoamb, knowtheydontknow}. 

Several studies have explored methods for resolving ambiguities in questions, often targeting issues like missing qualifiers that can lead to different interpretations~\citep{alignamb, resolveamb}. Many research on questions outside the model's knowledge scope has focused on guiding models to recognize when they lack sufficient information to answer, thereby reducing hallucinations~\citep{knowtheydontknow, hallu1, hallu2}. However, a commonly overlooked real-world scenario involves individuals posing seemingly relevant questions that are unanswerable due to their limited familiarity with the document's knowledge domain. In such cases, large language models can be employed to reformulate these questions, enabling the inquirers to obtain the information they likely intended to seek.

The issue of document-related unanswerable questions was first systematically addressed in a dataset by~\citet{unans1}, and later expanded by studies such as~\citet{yuerrorrate}, which analyzed Google search queries, and~\citet{kimdata}, which examined Reddit discussions. Both studies focused on identifying questions with incorrect assumptions or presuppositions that made them unanswerable.

In recent work,~\citet{couldask} introduced a high-quality dataset of document-related unanswerable questions, exploring how large language models can assist in reformulating such questions. They evaluated several well-known large language models using basic approaches like zero-shot and more advanced ones such as few-shot or Chain of Thought (CoT) prompting~\citep{cot}, but the results were suboptimal. In this paper, we propose the zero-shot DRS method to significantly improve LLMs' ability to help users reformulate unanswerable questions, advancing progress in this area.

\section{Methodology}
In this paper, we propose a new zero-shot method called DRS: \textbf{D}eep Question \textbf{R}eformulation with \textbf{S}tructured Output, which significantly improves LLMs' ability to reformulate unanswerable questions based on a given document, helping users obtain desired answers in unfamiliar knowledge domains. 

Our DRS method combines DFS (Depth-First Search) algorithm with LLMs to address challenges posed by human-generated unanswerable questions involving multiple entities. Directly inputting all entities into LLMs for question reformulation risks overlooking the lack of meaningful relationships between them, leading to unreliable or incoherent answers. To mitigate this, our DFS-based approach systematically explores semantically related entity combinations, ensuring the generation of meaningful questions. Additionally, by controlling the search depth and iteration count, our DRS method strikes a balance between computational efficiency and accuracy, yielding high-quality results with minimal overhead.

Our approach consists of three main steps: (i) Entity extraction and filtering. (ii) DFS combination search and structured question generation. (iii) Candidate question re-evaluation. The detailed process of our zero-shot DRS method is illustrated in Figure \ref{fig:algorithm}.

\subsection{Entity Extraction and Filtering}
When people encounter a text with unfamiliar knowledge, they usually raise questions about the key entities in the text, indicating their desire to understand more about these entities. Therefore, when reconstructing questions, we focus on the key entities from the original question. This ensures that the reconstructed question addresses the content people genuinely care about, rather than an arbitrary, overly simple new question.

For instance, consider the question: `When does Rainer Hertrich, the German co-head of EADS, step down?'. The key entities in this example are `Rainer Hertrich', `German', and `EADS', which represent the core subject and its relevant modifiers. While modern LLMs are capable of extracting important entities from a question, they sometimes include verb phrases such as `step down', which, although significant to the context, do not qualify as entities. This tendency can lead to the inclusion of extraneous elements, resulting in less precise entity identification and subsequently higher error rates in question reformulation task. 

To address this limitation, we add a classification framework that categorizes entities into five semantic roles based on their function within the question. This approach facilitates the identification of core entities that align with the user's intent while filtering out less critical elements, thereby reducing ambiguities in reformulation and enhancing overall precision.

Therefore, this process involves two following sub-steps:

(i) We use a simple and direct zero-shot prompt to have the large language model extract all entities it considers important from the original question. The goal is to minimize the omission of any potentially important entities.

(ii) We then apply the large language model again as an effective entity classifier to classify the previously extracted entities into five categories: subject, object, predicate, attribute, and others. We retain all entities classified as subject, object, and attribute, as these are the most important components of the question, and discard the others.

\begin{table*}[!tb]
	\centering
	\begin{adjustbox}{width=1.00\linewidth}
        \renewcommand{\arraystretch}{1.20}
        \setlength{\tabcolsep}{13pt}
        \resizebox{\linewidth}{!}{
    	\begin{tabular}{cccccc}
            \toprule
            \textbf{Subset Name} & \textbf{Test Data Size} & \textbf{Document Length} & \textbf{Question Length} & \textbf{Entity Numbers} & \textbf{Domain} \\ \midrule \midrule
            SQuADv2 & $507$ / $1000$ & $189.74 \pm 77.59$ & $13.85 \pm 4.32$ & $2.29 \pm 0.95$ & Wikipedia \\
            QA\textsuperscript{2} & 247 / 506 & $1109.45 \pm 899.83$ & $10.29 \pm 2.37$ & $2.05 \pm 0.57$ & Mostly Wikipedia \\
            BanditQA & $736$ / $2070$ & $363.67 \pm 202.03$ & $10.36 \pm 3.16$ & $1.70 \pm 0.72$ & Wikipedia \\
            BBC & $59$ / $278$ & $652.90 \pm 385.85$ & $20.61 \pm 4.73$ & $3.31 \pm 1.04$ & News \\
            Reddit & $113$ / $313$ & $569.48 \pm 364.64$ & $16.58 \pm 3.49$ & $2.93 \pm 0.78$ & Social Media \\
            Yelp & $51$ / $165$ & $486.87 \pm 184.44$ & $17.72 \pm 3.82$ & $3.04 \pm 0.76$ & Review \\ \bottomrule
            \end{tabular}
            \vspace{-2mm}
        }
	\end{adjustbox}
	\caption{The detail information of all experimental data for evaluating the performance on the question reformulation task. The \textit{Length} is calculated based on the number of tokens after tokenization using \gemmatwonineb.
    \label{tab:datainfo}}
\end{table*}

\subsection{DFS Combination Search and Structured Question Generation}
After obtaining reliable entities, we proceed with question reformulation. We use the DFS-based algorithm to explore combinations of different entities. For any combination that could reformulate a new question, we allow it to generate a completely new question and store it for later evaluation. Specifically, in this step, we focus on the following processes:

(1) We use the DFS algorithm to select possible entity combinations. When the number of entities in a combination exceeds half of the filtered entities, we move on to step (2).

(2) We prompt the large language model with a concise zero-shot instruction to generate a structured statement containing all selected entities, based on the chosen entities and corresponding text. This statement serves as a conclusion that can be drawn from the document.

(3) After generating the statement, we ask the large language model to create a structured question that includes all the selected entities and can be answered, based on the generated statement.

(4) We then return the generated question to the large language model and ask it to verify whether the question includes all the necessary entities. If it does, we proceed to the next step (5); if not, we return to step (1) and select a new combination.

(5) If the question contains all the entities, we input it again into the large language model to check whether it can be answered. If it can, we store both the question and the count of its entities for final selection. If not, we discard it and return to step (1) to search for another combination. (Note: When the number of valid questions reaches the set threshold, no further combinations will be attempted.)

In this step, our algorithm uses a DFS search approach to identify possible entity combinations that can be reformulated into new questions. During the generation process, we find that directly prompting the large language model to reshape the question often results in a low success rate due to the model's potentially unlimited output content. To address this, we guide the large language model to generate structured outputs by restricting the entities required in the generated text. We ensure the reliability of the reformulated question by first generating a statement, then using that statement to create the question. 

Additionally, to enhance the algorithm’s efficiency, we verify the entities immediately after generating the new question, preventing unnecessary subsequent steps. We also apply pruning to the DFS algorithm by limiting the total number of preselected questions and controlling the search depth.

\subsection{Candidate Question Re-evaluation}
The DFS search process generates multiple candidate questions, each preserving different aspects of the original question. To select the optimal reformulation, we develop a two-stage re-evaluation strategy.

First, we assess each candidate's answerability by prompting the large language model to verify whether the question can be answered using only the information present in the document. This ensures that our final selection maintains semantic validity and practical utility.

Second, we consider the entity overlap score of each answerable candidate. Questions with higher entity overlap are preferred as they better maintain the user's original intent. This is quantified by:

\[
\text{Entity Overlap Score} = \frac{\lvert \text{Entities}_{\text{cand}} \cap \text{Entities}_{\text{orig}} \rvert}{\lvert \text{Entities}_{\text{orig}} \rvert}
\]

When multiple candidates achieve similar answerability, we select the one with the highest entity overlap score. As the example illustrated in Figure \ref{fig:algorithm}, we identify three questions that are considered potentially answerable by the large language model. Therefore, we allow the model to select the most optimal reformulated question. While both Q2 and Q4 are recognized as clearly answerable, Q4 is ultimately chosen due to its inclusion of a greater number of valid entities. Thus, the model returns Q4 as the final reformulated question to the user. This approach ensures that the final output: (1) can be answered from the given document, (2) maximally preserves the user's original intent, and (3) maintains semantic coherence with the source text.

\section{Experiments}
\subsection{Datasets}
In this paper, we utilize the newly constructed high-quality dataset \textbf{CouldAsk}\footnote{\url{https://huggingface.co/datasets/wentingzhao/couldask}}, introduced by~\citet{couldask}. This dataset comprises subsets from six diverse sources, including Yelp, BBC, SQuAD, Reddit, and others, featuring over $4,000$ high-difficulty, high-quality data points filtered by both human annotators and \gptfour~\citep{gpt4}. For our study, we focus on over $1,700$ unanswerable questions selected from all subsets as test data for model reformulation. The detailed information about our experimental data is provided in Table \ref{tab:datainfo}.

\begin{table*}[!tb]
	\centering
	\begin{adjustbox}{width=1.00\linewidth}
        \renewcommand{\arraystretch}{1.25}
        \setlength{\tabcolsep}{10pt}
        \resizebox{\linewidth}{!}{
    	\begin{tabular}{lccccccc}
            \toprule
            \multirow{2}{*}{\textbf{Evaluator}} & \textbf{QA\textsuperscript{2}} & \textbf{BanditQA} & \textbf{BBC} & \textbf{Reddit} & \textbf{Yelp} & \textbf{SQuADv2} & \textbf{Average} \\ \cmidrule{2-8} 
             & $200$ / $247$ & $300$ / $507$ & $50$ / $59$ & $100$ / $113$ & $50$ / $51$ & $300$ / $736$ & $1000$ / $1713$ \\ \midrule \midrule
            $\llamatwosevenb$~\citep{couldask} & $39.00$ & $68.67$ & $14.00$ & $79.00$ & $44.00$ & $72.00$ & $52.78$ \\ \midrule
            $\gptfouromini$ & $90.50$ & $92.33$ & $88.00$ & $92.00$ & $90.00$ & $91.33$ & $90.70$ \\ \bottomrule
            \end{tabular}
            \vspace{-2mm}
        }
	\end{adjustbox}
	\caption{The comparison of prediction accuracy between previous \llamatwosevenb evaluator proposed by~\citet{couldask} and our proposed \gptfouromini evaluator.
    \label{tab:evaluatorcomparison}}
\end{table*}

\subsection{Large Language Models}
To evaluate our methodology and compare it with baseline approaches from previous studies, we conduct all experiments using four different LLMs: \gptthreepointfive, \gptfouromini, \gemmatwonineb and \qwenseven~\citep{gemma2, qwen2.5}. The detailed information of different LLMs is shown in Appendix \ref{sec:llmintro}.

\subsection{Metric}
We employ the metric \textbf{Accuracy}, in line with previous research, to measure the proportion of unanswerable questions successfully reformulated. This evaluation involves two criteria:

(i) We use \gptfouromini as an evaluator to generate reasoning steps and assess whether the reformulated question can be answered using the relevant text. (ii) We use \gptfouromini as an entity detector to measure the overlap of entities between the reformulated question and the original question. If the number of overlapping entities exceeds half of the labeled entities in the dataset, we consider the reformulated question likely to help the inquirer obtain the desired information. 

A reformulated question is deemed successful and counted towards accuracy only if it satisfies both criteria.

\section{GPT Evaluator}
In the previous paper,~\citet{couldask} fine-tuned the \llamatwosevenb model~\citep{llama2} on the training dataset for the sequence classification task to determine whether a question could be answered from the corresponding text. This approach achieved high scores on their validation dataset. However, in our experiments, we find that its generalization performance is suboptimal; even on the \textbf{CouldAsk} dataset, which includes training data, it still yields extremely poor classification results.

% \footnote{\texttt{Input Format: \{context\}\textbackslash n\{question\}}}

Given the situation described above, we find it necessary to propose a reliable and efficient evaluation method. Therefore, in this paper, we choose the \gptfouromini model, which demonstrates excellent language understanding capabilities, as the evaluation model, ensuring both strong evaluation performance and cost-effectiveness. 

To demonstrate the strong capabilities of the \gptfouromini model, we select a subset of labeled questions from each portion of the \textbf{CouldAsk} dataset. We then have both evaluation models perform classification predictions and record their classification accuracy on this data, with the results shown in Table \ref{tab:evaluatorcomparison}. 

We observe that \gptfouromini consistently achieves high scores across all tests, whereas the fine-tuned \llamatwosevenb model shows significantly lower accuracy than \gptfouromini and exhibits substantial score variation across different datasets (ranging from a high of $79\%$ to a low of $14\%$). For the average accuracy across six different datasets, our \gptfouromini evaluator achieves over $90\%$, while the \llamatwosevenb evaluator only reaches $52\%$, performing slightly better than random predictions. This clearly indicates that the fine-tuned model lacks generalizability across diverse data. Therefore, using \gptfouromini as the evaluator is not only more reliable but also provides faster evaluation and lower operational costs.

\begin{table*}[!tb]
	\centering
	\begin{adjustbox}{width=1.00\linewidth}
        \renewcommand{\arraystretch}{1.25}
        \setlength{\tabcolsep}{10pt}
        \resizebox{\linewidth}{!}{
    	\begin{tabular}{llccccccc}
            \toprule
            \multicolumn{1}{l}{\textbf{Model}} & \multicolumn{1}{l}{\textbf{Method}} & \textbf{\,\,\,QA\textsuperscript{2}\,\,\,} & \textbf{BanditQA} & \textbf{\,\,\,\,BBC\,\,\,\,} & \textbf{\,\,\,Reddit\,\,\,} & \textbf{\,\,\,\,Yelp\,\,\,\,} & \textbf{SQuADv2} & \textbf{Average} \\ \midrule \midrule
            \multirow{5}{*}{\gptthreepointfive} & $\texttt{Zero-Shot}$ & $28.74$ & $11.41$ & $13.56$ & $18.58$ & $19.61$ & $36.69$ & $21.43$ \\
             & $\texttt{Zero-Shot w/ CoT}$ & $29.15$ & $15.63$ & $13.56$ & $20.35$ & $15.69$ & $43.79$ & $23.03$ \\
             & $\texttt{Few-Shot}$ & $32.79$ & $22.69$ & $15.25$ & $23.01$ & $29.41$ & $38.46$ & $26.94$ \\
             & $\texttt{Few-Shot w/ CoT}$ & $44.94$ & $48.78$ & $16.95$ & $18.58$ & $25.49$ & $41.22$ & $32.66$ \\
             & $\texttt{\textbf{Zero-Shot DRS (ours)}}$ & $\textbf{81.80}$ & $\textbf{73.20}$ & $\textbf{62.71}$ & $\textbf{75.22}$ & $\textbf{66.67}$ & $\textbf{62.90}$ & $\textbf{70.42}$ \\ \midrule
            \multirow{5}{*}{\gptfouromini} & $\texttt{Zero-Shot}$ & $36.44$ & $15.08$ & $15.25$ & $23.01$ & $15.69$ & $47.93$ & $25.57$ \\
             & $\texttt{Zero-Shot w/ CoT}$ & $49.39$ & $37.50$ & $42.37$ & $17.70$ & $35.29$ & $50.10$ & $38.73$ \\
             & $\texttt{Few-Shot}$ & $36.44$ & $17.80$ & $16.95$ & $21.24$ & $21.57$ & $46.55$ & $26.76$ \\
             & $\texttt{Few-Shot w/ CoT}$ & $61.13$ & $52.85$ & $47.46$ & $43.36$ & $56.86$ & $59.57$ & $53.54$ \\
             & $\texttt{\textbf{Zero-Shot DRS (ours)}}$ & $\textbf{88.26}$ & $\textbf{80.16}$ & $\textbf{79.66}$ & $\textbf{83.19}$ & $\textbf{78.43}$ & $\textbf{78.30}$ & $\textbf{81.33}$ \\ \midrule
            \multirow{5}{*}{\qwenseven} & $\texttt{Zero-Shot}$ & $46.15$ & $24.05$ & $25.42$ & $23.89$ & $25.49$ & $46.70$ & $31.95$ \\
             & $\texttt{Zero-Shot w/ CoT}$ & $47.18$ & $44.21$ & $27.12$ & $35.40$ & $31.37$ & $53.55$ & $39.81$ \\
             & $\texttt{Few-Shot}$ & $39.68$ & $25.14$ & $15.25$ & $23.89$ & $19.61$ & $39.64$ & $27.20$ \\
             & $\texttt{Few-Shot w/ CoT}$ & $54.28$ & $45.59$ & $25.42$ & $25.66$ & $33.33$ & $50.64$ & $39.15$ \\
             & $\texttt{\textbf{Zero-Shot DRS (ours)}}$ & $\textbf{77.73}$ & $\textbf{69.84}$ & $\textbf{55.93}$ & $\textbf{53.10}$ & $\textbf{58.82}$ & $\textbf{60.55}$ & $\textbf{62.66}$ \\ \midrule
            \multirow{5}{*}{\gemmatwonineb} & $\texttt{Zero-Shot}$ & $36.03$ & $21.33$ & $20.34$ & $18.58$ & $19.61$ & $42.21$ & $26.35$ \\
             & $\texttt{Zero-Shot w/ CoT}$ & $29.15$ & $17.12$ & $22.03$ & $23.01$ & $19.61$ & $32.94$ & $23.96$ \\
             & $\texttt{Few-Shot}$ & $26.32$ & $16.17$ & $25.42$ & $15.04$ & $17.65$ & $29.39$ & $21.67$ \\
             & $\texttt{Few-Shot w/ CoT}$ & $22.67$ & $18.07$ & $10.17$ & $24.78$ & $25.49$ & $27.42$ & $21.43$ \\
             & $\texttt{\textbf{Zero-Shot DRS (ours)}}$ & $\textbf{59.92}$ & $\textbf{60.73}$ & $\textbf{55.93}$ & $\textbf{59.29}$ & $\textbf{49.02}$ & $\textbf{55.62}$ & $\textbf{56.75}$ \\ \bottomrule
            \end{tabular}
            \vspace{-2mm}
        }
	\end{adjustbox}
	\caption{The main experimental results of four different large language models on full data across six different datasets, where best results are highlighted in \textbf{bold} font.
    \label{tab:mainexp}}
\end{table*}

\section{Experiment Results}
\begin{figure*}[!tb]
    \centering
    \includegraphics[width=1.00\linewidth]{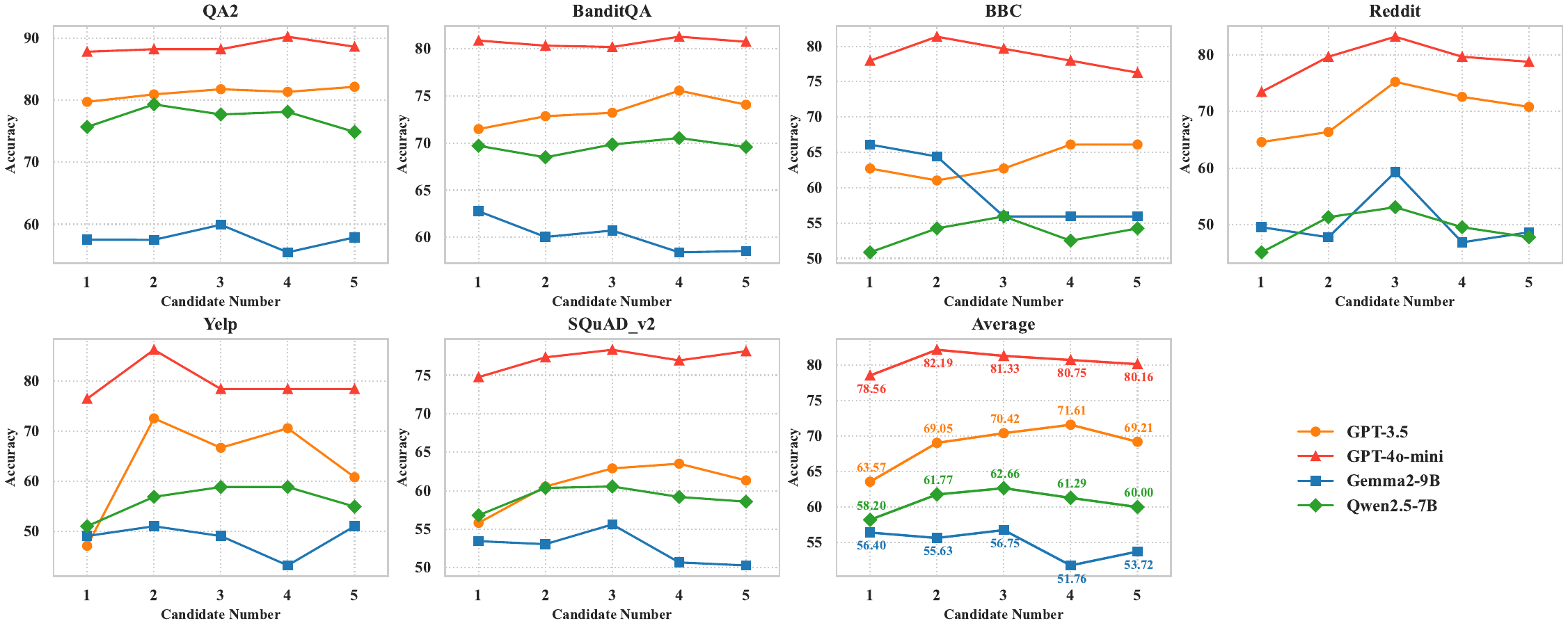}
    \caption{Model accuracy across four large language models with varying numbers of candidate questions, evaluated on six datasets and their average.
    \label{fig:accnumcandidate}}
    \vspace{-2mm}
\end{figure*}

\begin{figure}[!tb]
    \centering
    \includegraphics[width=1.00\linewidth]{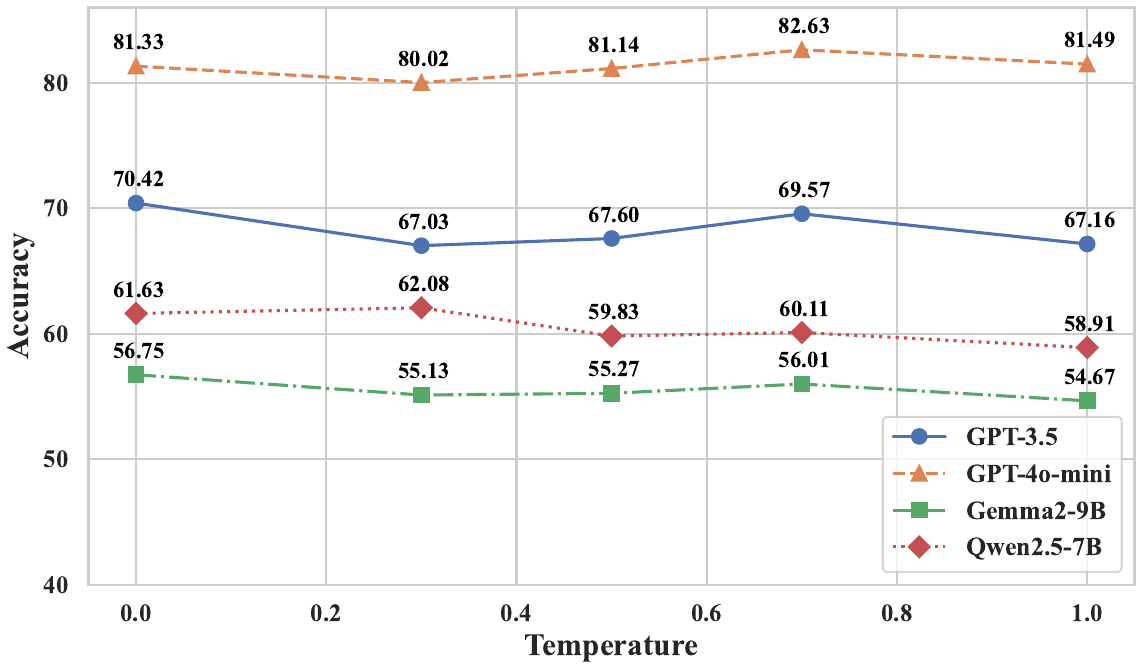}
    \caption{Model accuracy across four large language models with five different temperature settings, evaluated on six datasets and shows their average.
    \label{fig:tempcompare}}
    \vspace{-2mm}
\end{figure}

To demonstrate the effectiveness of our proposed zero-shot DRS method, we test it on six datasets, totaling over $1,700$ data points, using four different large language models. 
We conduct a fair and thorough comparison between our method and four baseline methods from previous research, with all accuracy scores presented in Table \ref{tab:mainexp}. We incorporate the phrase `think step by step' to elicit the models' chain-of-thought reasoning capabilities~\citep{thinkstepbystep, cot}.
Additionally, we adjust relevant parameters, such as temperature and the number of candidate questions, in the zero-shot DRS method and conduct multiple experiments. The results are shown in Figure \ref{fig:tempcompare} and Figure \ref{fig:accnumcandidate}. Finally, considering GPU resources and API costs, we select a subset of test data for additional experiments with more models, as listed in Appendix \ref{sec:moreexp}.

\stitle{Effective DRS}
Table \ref{tab:mainexp} clearly demonstrates the strong capabilities of our proposed zero-shot DRS method. Our method significantly outperforms all baselines across all datasets and models. Using the \gptthreepointfive model, the average accuracy in the zero-shot setting increases from $23\%$ to $70\%$, nearly tripling the original performance, which is a remarkable improvement that underscores the effectiveness of our approach. With the \gemmatwonineb model, while the results are lower than those of \gptthreepointfive, our method still boosts the zero-shot accuracy from $26\%$ to $57\%$, achieving an improvement of over $100\%$. Moreover, even when compared to few-shot baselines, our zero-shot method consistently outperforms them. On \gptthreepointfive, the $70\%$ average accuracy is more than double the $33\%$ accuracy of the few-shot CoT method. For \gemmatwonineb, the few-shot accuracy is even lower than the zero-shot baseline.

In addition to the significant improvements observed with \gemmatwonineb and \gptthreepointfive, both \gptfouromini and \qwenseven also show remarkable enhancements. For \gptfouromini, the accuracy achieved by zero-shot DRS is more than double that of zero-shot CoT, and still shows a $52\%$ improvement over the few-shot CoT results. While \qwenseven exhibits the smallest improvement ratio among the four models, it still achieves over a $57\%$ enhancement compared to both zero-shot and few-shot CoT. These comparisons collectively demonstrate the effectiveness of the zero-shot DRS method in question reformulation.

\stitle{Robust DRS}
In the process of our DRS method, several steps involve using LLMs to generate corresponding answers. The model's temperature influences the output, affecting the diversity and reliability of the responses. Due to the continuity and interrelation of these steps, errors can propagate and accumulate. Therefore, we apply different temperatures to various LLMs in our method, and the experimental results are presented in Figure \ref{fig:tempcompare}.

We observe that the average scores remain remarkably stable as the temperature increases. The score differences between different temperature settings are at most within 3 percentage points, which is negligible compared to the performance improvement brought by our method. Additionally, we notice that \gptthreepointfive and \gptfouromini achieve better scores when the temperature is set to $0.0$ or $0.7$. This is likely because, at a temperature of $0.0$, the models can more strictly follow the instructions containing entities, while at $0.7$, the diversity of the outputs is effectively utilized, providing more varied options for candidate questions and increasing the likelihood of selecting a better final question. In contrast, the accuracy of \qwenseven and \gemmatwonineb slightly decreases with higher temperatures, as smaller models struggle to follow instructions effectively as the temperature increases.

\stitle{Impact of the Number of Candidate Questions}
In our DRS method, before returning the final output question, we guide the model to generate a certain number of candidate questions. To explore the impact of this parameter on the performance of our method, we conduct experiments with varying numbers of candidate questions and observe how accuracy changes. We set the number of candidate questions from 1 to $5$ for each large language model and test on all datasets, with the results shown in Figure \ref{fig:accnumcandidate}. 

We observe that the number of candidate questions does affect the final accuracy. Notably, when only one candidate question is generated, the accuracy of question reformulation across various models drops more significantly compared to other settings. For instance, with \gptthreepointfive, setting the number of candidate questions to $1$ results in a decrease in accuracy from $70.42\%$ to $63.57\%$, a drop of nearly $10\%$. Similarly, \gptfouromini’s accuracy decreases from $81.33\%$ to $78.56\%$. This is because, when only one question is generated, inaccuracies in entity extraction, coupled with a shallow search depth in our DRS algorithm, make it challenging to ensure significant overlap between the entities in the reformulated question and the true label entities. Consequently, the reformulated question often fails to meet the entity overlap requirement.

\begin{figure*}[!tb]
    \centering
    \includegraphics[width=1.00\linewidth]{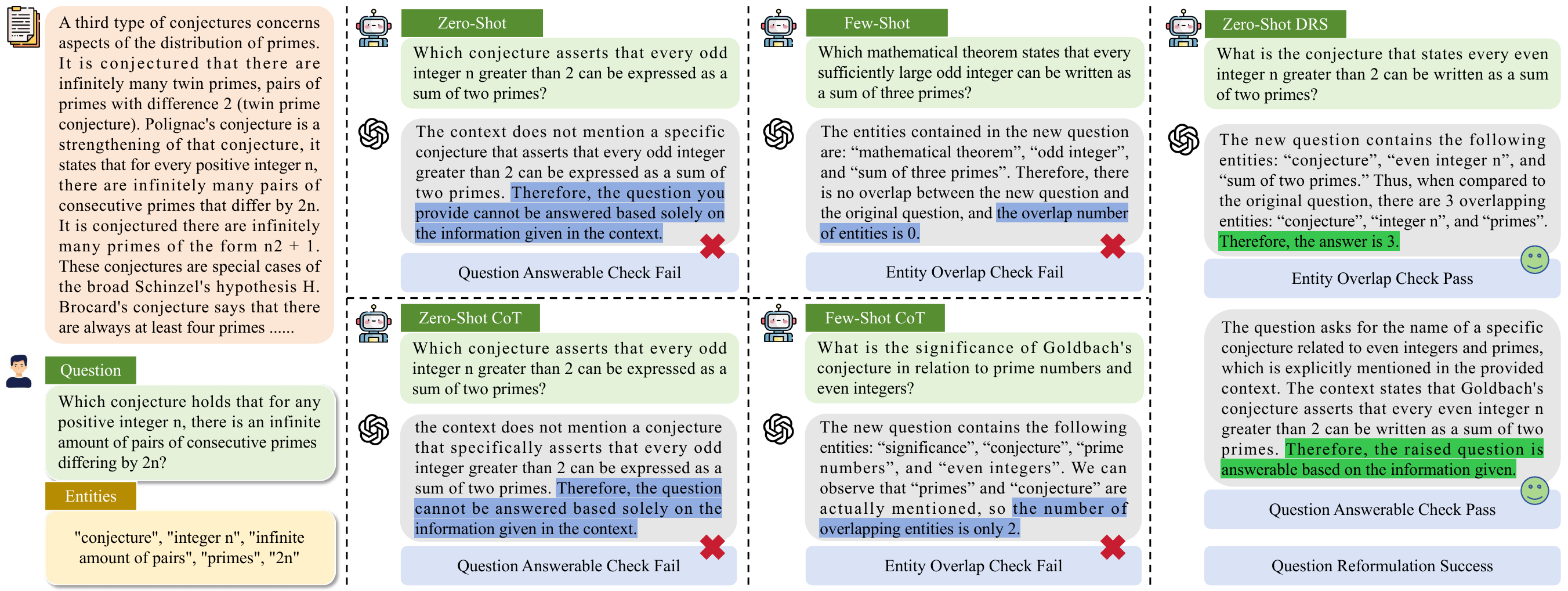}
    \caption{Comparison of our zero-shot DRS method with four baseline approaches, where our method successfully reformulates the question while all other baselines do not.
    \label{fig:caseshow}}
    \vspace{-2mm}
\end{figure*}

However, when the number of candidate questions is set from $2$ to $5$, the accuracy differences are relatively minor, typically within $3$ percentage points (only \qwenseven shows a larger decrease when candidate number is $4$), following a general trend of initially increasing and then decreasing. Besides, the curve shows that setting the number of candidate questions to $2$ or $3$ yields the most ideal results, achieving high accuracy while significantly saving search time. Even more convincing is the fact that, even when the number of candidate questions is set to the least ideal value of $1$, the scores obtained by our DRS method still far outperform all other baselines. This strongly demonstrates the power and remarkable robustness of our proposed method.

\section{Case Study}
To highlight the advantages of our proposed zero-shot DRS method, we select some examples to demonstrate its effectiveness compared to other baselines, as shown in Figure \ref{fig:caseshow}. We observe that the relevant documents primarily discuss mathematical theorems, and the user's question involves five entities but cannot be answered using the corresponding documents. In these examples, the reformulated questions generated using zero-shot and zero-shot CoT are identical. These questions overlap with the original question on three entities: `conjecture', `integer n', and `primes'. However, they still cannot be answered based on the provided documents, making such reformulations unsuccessful.

In contrast, questions generated using the few-shot or few-shot CoT methods are more diverse. However, despite being answerable based on the document, they fail to pass the entity overlap check. Specifically, the questions generated by the few-shot method contain no entities from the original question, while those generated by the few-shot CoT method include only two entities. Both fail to meet the required entity overlap ratio of $0.5$ or higher, making these reformulations unsatisfactory.

In comparison, the reformulated question generated by our proposed DRS method satisfies both strict conditions: answerability and entity overlap ratio. The DRS method generates questions through a DFS search process, where the model is guided by the constraints of three entities: `conjecture', `integer n', and `primes'. In this process, we first prompt the model to construct a statement based on the document, using these entities. We then use this statement to generate a reformulated question containing the same entities. This approach ensures both answerability and entity overlap, optimizing the model's ability to help users obtain relevant answers.

From these observations, we conclude that even advanced methods like few-shot or few-shot CoT struggle to produce satisfactory reformulations due to the lack of constraints on generated content. This lack of control often results in outputs that either overly focus on preserving the original question’s entities, leading to unanswerable questions, or prioritize answerability at the expense of entity overlap. In contrast, the DRS method effectively balances these two requirements by systematically searching for entity combinations and generating structured statements and reformulated questions containing the specified entities. This approach unlocks the full potential of LLMs, enabling more effective question rephrasing for users.

\section{Conclusion}
In our paper, we propose a zero-shot DRS method that significantly enhances the performance of large language models on the unanswerable question reformulation task. 
Our method outperforms all other baselines, including the few-shot CoT approach, across six different datasets and various LLMs. We also conduct extensive experiments on key parameters, such as temperature and the number of candidate questions, with results demonstrating the strong capabilities and robustness of our approach. 
Furthermore, our findings reveal that the previously proposed \llamatwosevenb evaluator lacks the capability to fairly assess the reformulated questions. We propose using \gptfouromini as a more reliable, accurate, and low-latency evaluator for future research. 
Looking ahead, we aim to develop even more effective methods to harness the full potential of LLMs, helping them assist people in understanding unfamiliar documents more effectively.

\section*{Limitations}
In this paper, our proposed zero-shot DRS method significantly enhances the ability of LLMs to reformulate unanswerable questions. However, there are still the following limitations: (i) Despite the use of efficient pruning with the DFS algorithm, the process still requires multiple passes through the document, which increases computational costs. Future research could explore methods to enable LLMs to complete the reformulation in a single attempt. (ii) While we conduct tests on six datasets, they do not include documents from specific disciplines, which are common in real-world applications. Future work could focus on developing datasets that cover a broader range of domains, allowing for a more comprehensive evaluation of algorithmic and large language model capabilities.

\section*{Ethics Statement}
Ethical considerations are of utmost importance in our research endeavors. In this paper, we strictly adhere to ethical principles by exclusively utilizing open-source datasets and employing various models that are either open-source or widely recognized in the scientific community. Our proposed methodology aims to enhance the model's ability to reformulate unanswerable questions when encountering documents from new knowledge domains in real-world scenarios. We are committed to upholding ethical standards throughout the research process, prioritizing transparency, and promoting the responsible use of technology for the betterment of society. 

\section*{Acknowledgments}
This research is supported by Optum Labs, DARPA ANSR program FA8750-23-2-0004, a National Science Foundation CAREER award \#2339766, and University of California, Merced.
% The views and conclusions are those of the authors and should not reflect the official policy or position of DARPA or the U.S. Government.

\bibliography{custom}
\bibliographystyle{acl_natbib}

\newpage
\appendix

\section{Implementation Details}
\label{sec:impdetails}
We use the \textit{gpt-3.5-turbo-0125} version for \gptthreepointfive, \textit{gpt-4-0125-preview} for \gptfour, and \textit{gpt-4o-mini} for \gptfouromini. 
All experiments were conducted between November 9th and 25th, 2024.

To ensure fair experimental comparisons and re-evaluate the baseline scores reported in the previous paper, we use the open-source GitHub code provided by~\citet{couldask}, maintaining consistency in the prompts and LLM parameters. 
The versions of all OpenAI models also remain consistent with the ones mentioned above.

In our main experiments using the proposed DRS method, we set the temperature to $0.0$ for all large language models to ensure reproducibility. Similarly, for the \gptfouromini evaluator, we set the \textit{temperature} to $0.0$ and \textit{top\_p} to $1.0$ to maintain reliability and consistency in the model's judgments. For the two \gpt series models, we use the official OpenAI API\footnote{\url{https://openai.com/}} for inference. 

For the two open-source models \gemmatwonineb and \qwenseven, we use the weights released on HuggingFace\footnote{\url{https://huggingface.co/}} and deploy them on a single NVIDIA RTX A6000 GPU using \textbf{bfloat16} precision and utilize the generate function for text generation.
% (Additional details are provided in Appendix \ref{sec:impdetails}).

\section{Large Language Models}
\label{sec:llmintro}
\stitle{Closed-Source Models} To balance experimental cost and effectiveness, we primarily use two widely adopted models from OpenAI: \gptthreepointfive and \gptfouromini. In Appendix \ref{sec:moreexp}, we select a subset for the experiment using \gptfour.

\begin{itemize}
    \item {\gptthreepointfive}: A robust large language model developed by OpenAI, capable of generating text based on instructions, and highly effective across diverse natural language processing tasks.
    \item {\gptfour}: An advanced multi-modal language model from OpenAI that accepts both image and text inputs for text generation, achieving near-human performance on various benchmarks.
    \item {\gptfouromini}: A cost-efficient multi-modal model released by OpenAI on July 18, 2024, offering low latency and cost while supporting a wide range of tasks.
\end{itemize}

\stitle{Open-Source Models} We also perform experiments using two well-known models developed by Google and Alibaba: \gemmatwonineb~\citep{gemma2} and \qwenseven~\citep{qwen2, qwen2.5}. In Appendix \ref{sec:moreexp}, we select a subset for the experiment using \llamaseventy and \gemmatwotwosevenb.

\begin{itemize}
\item {\llamaseventy}: The $70$B-parameter flagship model from the latest \llamathreepointone open-source family released by MetaAI.
\item {\gemmatwo}: The next-generation open-source model from Google, released on June 27, 2024, as an improved version of {\gemma}, available in $2$B, $9$B, and $27$B parameter configurations.
\item {\qwenseven}: A $7$B-parameter model from Alibaba’s latest \qwentwopointfive series, delivering strong performance on various benchmarks compared to other models of similar size.
\end{itemize}

\begin{table*}[!tb]
	\centering
	\begin{adjustbox}{width=1.00\linewidth}
        \renewcommand{\arraystretch}{1.40}
        \setlength{\tabcolsep}{6pt}
        \resizebox{\linewidth}{!}{
    	\begin{tabular}{llccccccc}
            \toprule
            \multicolumn{1}{l}{\textbf{Model}} & \multicolumn{1}{l}{\textbf{Method}} & \textbf{\,\,\,QA\textsuperscript{2}\,\,\,} & \textbf{BanditQA} & \textbf{\,\,\,\,BBC\,\,\,\,} & \textbf{Reddit} & \textbf{\,\,\,\,Yelp\,\,\,\,} & \textbf{SQuADv2} & \textbf{Average} \\ \midrule \midrule
            \multirow{3}{*}{\gptfour} & $\texttt{Zero-Shot}$ & $44.00$ & $30.00$ & $28.81$ & $25.66$ & $21.57$ & $64.00$ & $35.67$ \\
             & $\texttt{Few-Shot}$ & $48.00$ & $29.33$ & $37.29$ & $23.01$ & $35.29$ & $60.67$ & $38.93$ \\
             & $\texttt{\textbf{Zero-Shot DRS (ours)}}$ & $\textbf{78.67}$ & $\textbf{72.67}$ & $\textbf{71.19}$ & $\textbf{66.37}$ & $\textbf{62.75}$ & $\textbf{70.00}$ & $\textbf{70.28}$ \\ \midrule
             \multirow{3}{*}{\llamaseventy} & $\texttt{Zero-Shot}$ & $36.00$ & $31.33$ & $33.90$ & $19.47$ & $21.58$ & $58.00$ & $33.38$ \\
             & $\texttt{Few-Shot}$ & $29.33$ & $29.33$ & $27.12$ & $18.58$ & $25.49$ & $36.67$ & $27.75$ \\
             & $\texttt{\textbf{Zero-Shot DRS (ours)}}$ & $\textbf{65.33}$ & $\textbf{74.67}$ & $\textbf{67.80}$ & $\textbf{61.95}$ & $\textbf{74.51}$ & $\textbf{66.67}$ & $\textbf{68.48}$ \\ \midrule
             \multirow{3}{*}{\gemmatwotwosevenb} & $\texttt{Zero-Shot}$ & $10.00$ & $12.00$ & $23.73$ & $18.58$ & $17.65$ & $42.67$ & $20.77$ \\
             & $\texttt{Few-Shot}$ & $28.67$ & $8.00$ & $6.78$ & $4.42$ & $13.73$ & $20.67$ & $13.71$ \\
             & $\texttt{\textbf{Zero-Shot DRS (ours)}}$ & $\textbf{42.00}$ & $\textbf{64.00}$ & $\textbf{52.54}$ & $\textbf{46.02}$ & $\textbf{37.25}$ & $\textbf{55.33}$ & $\textbf{49.52}$ \\ \bottomrule
            \end{tabular}
            \vspace{-2mm}
        }
	\end{adjustbox}
	\caption{The additional experimental results of three different large language models on full data across six different datasets, where best results are highlighted in \textbf{bold} font.
    \label{tab:addexp}}
\end{table*}

\section{Additional Experiments}
\label{sec:moreexp}
To address the constraints of GPU resources and API costs, we randomly select appropriately sized subsets from each test dataset for experiments with three well-known models: \gptfour, \llamaseventy, and \gemmatwotwosevenb, which demonstrates the generalizability of our approach. We use the complete datasets for Yelp, BBC, and Reddit, while for the three larger datasets, we randomly select $150$ samples for the experiments. For comparison, we focus on the results of our DRS method against two representative baselines, zero-shot and few-shot, due to the high computational cost of the models. The experimental results appear in Table \ref{tab:addexp}.

Based on the experimental results, our proposed DRS method demonstrates significant improvements over both zero-shot and few-shot baselines across three newly tested large language models. For instance, on \llamaseventy, zero-shot DRS achieves over a $100\%$ improvement in accuracy compared to the baselines. Similarly, on \gptfour, the improvement approaches $100\%$. Furthermore, with the DRS method, the average reformulation success rate for these two models reaches nearly $70\%$, which is remarkable. In contrast, \gemmatwotwosevenb achieves an average accuracy of only about $50\%$ with DRS, lagging behind the other two models. This discrepancy may be attributed to differences in model size and instruction adherence. Nevertheless, on \gemmatwotwosevenb, DRS still boosts accuracy by approximately $125\%$ over the zero-shot baseline and by around $275\%$ over the few-shot baseline. Overall, these results further validate the effectiveness of our DRS method, highlighting its strong generalization and robustness across diverse large language models.

\begin{table}[!tb]
	\centering
	\begin{adjustbox}{width=1.00\linewidth}
        \renewcommand{\arraystretch}{1.50}
        \setlength{\tabcolsep}{10pt}
        \resizebox{\linewidth}{!}{
    	\begin{tabular}{llc}
            \toprule
            \textbf{Model} & \textbf{Method} & \textbf{Inference Time} \\ \midrule \midrule
            \multirow{4}{*}{\gptthreepointfive} & Zero Shot w/ CoT & $6.80s$ \\
             & Few Shot w/ CoT & $7.32s$ \\
             & DRS (candidate num = $2$) & $10.07s$ \\
             & DRS (candidate num = $3$) & $11.44s$ \\ \midrule
            \multirow{4}{*}{\gemmatwonineb} & Zero Shot w/ CoT & $12.51s$ \\
             & Few Shot w/ CoT & $13.82s$ \\
             & DRS (candidate num = $2$) & $17.36s$ \\
             & DRS (candidate num = $3$) & $19.65s$ \\ \bottomrule \bottomrule
            \end{tabular}
            \vspace{-2mm}
        }
	\end{adjustbox}
	\caption{Average inference time per sample across six datasets for three types of methods.
    \label{tab:inferencetime}}
\end{table}

\section{Inference Time of DRS} 
\label{sec:inferencetime}
In Table \ref{tab:inferencetime}, we present the average runtime of \gptthreepointfive and \gemmatwonineb under different methods. For our DRS method, we include the runtime when the number of candidate questions is set to $2$ or $3$, as Figure \ref{fig:accnumcandidate} shows that optimal performance can already be achieved with $2$ or $3$ candidates. From Table \ref{tab:inferencetime}, we observe that the DRS method, implemented based on a DFS search approach, results in longer runtimes compared to the few-shot CoT method, despite the use of effective pruning techniques. Specifically, when the number of candidate questions is $2$, the runtime increases by $37\%$ on \gptthreepointfive and $25\%$ on \gemmatwonineb. 

However, despite the slight increase in inference time, the accuracy improvements of the DRS method over the few-shot CoT method are remarkable, with increases of $113\%$ (from $32.66$ to $69.55$) and $160\%$ ($21.43$ to $55.63$) on the listed two large language models, respectively. This clearly shows a trade-off between runtime and accuracy, but the accuracy gains of the DRS method far outweigh the impact of the longer runtime. Therefore, we have strong evidence to support that the DRS method is a reasonable and efficient approach.

\begin{table}[!tb]
	\centering
	\begin{adjustbox}{width=1.00\linewidth}
        \renewcommand{\arraystretch}{1.30}
        \setlength{\tabcolsep}{40pt}
        \resizebox{\linewidth}{!}{
    	\begin{tabular}{|c|c|}
            \toprule
            \textbf{Dataset} & \textbf{Human Evaluation} \\ \midrule
            QA\textsuperscript{2} & $100\,/\,100\,(100\%)$ \\
            BanditQA & $100\,/\,100\,(100\%)$ \\
            BBC & $59\,/\,59\,(100\%)$ \\
            Reddit & $100\,/\,100\,(100\%)$ \\
            Yelp & $51\,/\,51\,(100\%)$ \\
            SQuADv2 & $99\,/\,100\,(99\%)$ \\ \bottomrule
            \end{tabular}
            \vspace{-2mm}
        }
	\end{adjustbox}
	\caption{Human evaluation results on six datasets for assessing the meaningfulness and relevance of reformulated questions.
        \label{tab:humaneval}}
\end{table}

\section{Human Evaluation}
In this paper, we propose a zero-shot DRS method to significantly improve the question reformulation task. However, a key concern arises: \textit{Do LLMs merely reformulate extremely simple questions that are unrelated to the passages but can be easily answered by the models themselves?} For example, \textit{How to spell wasabi?}

To thoroughly assess the effectiveness of the DRS method, we recruit three native English-speaking graduate students to evaluate all datasets. For subsets with over $100$ samples, we randomly selected $100$ reformulated questions for evaluation. A question is deemed valid if at least two students agree that it is meaningful and relevant to the passage. The human evaluation results are presented in Table \ref{tab:humaneval}.

The results clearly show that our proposed zero-shot DRS method generates reformulated questions that are nearly $100\%$ meaningful and highly relevant to the given long context. Only one reformulated question within the SQuADv2 subset is deemed unmeaningful by three evaluators. This highlights the effectiveness of our method in faithfully preserving the user's original intent and producing relevant reformulated questions.

\section{Experimental Prompts}
\label{sec:drsprompts}

\begin{figure}[!htbp]
    \centering
    \begin{tcolorbox}[top=5pt, bottom=5pt, colback=gray!10, boxrule=1pt, colframe=black, title=Entity Extraction Prompt, fonttitle=\fontsize{10}{0}\selectfont, fontupper=\fontsize{11}{15}\selectfont]
    Find out all entities in the following question: \greentext{\{question\}}. The entities you find must be exactly exist in the given question. You should only return the entities, separated by comma and space.
    \end{tcolorbox}
    \caption{Entity Extraction Prompt}
    \label{fig:entityextraction}
\end{figure}

\begin{figure}[!htbp]
    \centering
    \begin{tcolorbox}[top=5pt, bottom=5pt, colback=gray!10, boxrule=1pt, colframe=black, title=Entity Categorize Prompt, fonttitle=\fontsize{10}{0}\selectfont, fontupper=\fontsize{11}{15}\selectfont]
    Here is a question:\greentext{\{question\}}.
    
    Here is an entity in this question:\greentext{\{entity\}}.
    
    Tell me which category does this entity belong to in the given question - subject, object, predicate, attribute or others. You should only consider the situation of given entity in this specific question. Give your analysis within <analysis> tags, and only return its category name within <answer> tags.
    \end{tcolorbox}
    \caption{Entity Classification Prompt}
    \label{fig:entitycategorize}
\end{figure}

\begin{figure}[!htbp]
    \centering
    \begin{tcolorbox}[top=5pt, bottom=5pt, colback=gray!10, boxrule=1pt, colframe=black, title=Statement and Question Generation Prompt, fonttitle=\fontsize{10}{0}\selectfont, fontupper=\fontsize{11}{15}\selectfont]
    According to the following text: \greentext{\{context\}}.
    
    Generate a statement you could get from the given text that contains all following entities: \greentext{\{entities\}}. Then you are required to generate a question contains all given entities, which could be answered from your statement. Return the statement within <statement> tags and the question within <question> tags.
    \end{tcolorbox}
    \caption{Statement and Question Generation Prompt}
    \label{fig:questionreformulation}
\end{figure}

\begin{figure}[!htbp]
    \centering
    \begin{tcolorbox}[top=5pt, bottom=5pt, colback=gray!10, boxrule=1pt, colframe=black, title=Entity Overlap Evaluation Prompt, fonttitle=\fontsize{10}{0}\selectfont, fontupper=\fontsize{11}{15}\selectfont]
    Here is an original question: \greentext{\{question\}}, it contains the following entities: \greentext{\{entities\}}.
    
    Here is a new question: \greentext{\{new\_question\}}.
    
    Tell me the number of overlapping entities between the new question and the original question, they do not need to be strictly the same, as long as mentioned, uppercase or lowercase doesn't matter. Give your analysis within <analysis> tag, and only return the math number of overlap entities within <answer> tags.
    \end{tcolorbox}
    \caption{Entity Overlap Evaluation Prompt}
    \label{fig:entityoverlappingcalculate}
\end{figure}

\begin{figure}[!htbp]
    \centering
    \begin{tcolorbox}[top=5pt, bottom=5pt, colback=gray!10, boxrule=1pt, colframe=black, title=Question Answerability Evaluation Prompt, fonttitle=\fontsize{10}{0}\selectfont, fontupper=\fontsize{11}{15}\selectfont]
    Here is a long context: \greentext{\{context\}}.
    
    Here is a question: \greentext{\{question\}}.
    
    Tell me whether this question is answerable only according to the information in the provided context. Think carefully and give you analysis within <analysis> tags, then return only the final answer `yes' or `no' within <answer> tags.
    \end{tcolorbox}
    \caption{Question Answerability Evaluation Prompt}
    \label{fig:answerableevaluation}
\end{figure}

\end{document}